# Blockchain-Enabled Federated Learning Approach for Vehicular Networks

Shirin Sultana*, Jahin Hossain*, Maruf Billah*, Hasibul Hossain Shajeeb*, Saifur Rahman *,

Keyvan Ansari †  , and Khondokar Fida Hasan ‡  ,

*Department of Computer Science and Engineering, Bangladesh University of Business and Technology, Bangladesh

†School of Information Technology, Murdoch University, Australia

‡School of Professional Studies, University of New South Wales (UNSW), Australia Corresponding

Author Email: fida.hasan@unsw.edu.au

*Abstract*—Data from interconnected vehicles may contain sensitive information such as location, driving behavior, personal identifiers, etc. Without adequate safeguards, sharing this data jeopardizes data privacy and system security. The current centralized data-sharing paradigm in these systems raises particular concerns about data privacy. Recognizing these challenges, the shift towards decentralized interactions in technology, as echoed by the principles of Industry 5.0, becomes paramount. This work is closely aligned with these principles, emphasizing decentralized, human-centric, and secure technological interactions in an interconnected vehicular ecosystem. To embody this, we propose a practical approach that merges two emerging technologies: Federated Learning (FL) and Blockchain. The integration of these technologies enables the creation of a decentralized vehicular network. In this setting, vehicles can learn from each other without compromising privacy while also ensuring data integrity and accountability. Initial experiments show that compared to conventional decentralized federated learning techniques, our proposed approach significantly enhances the performance and security of vehicular networks. The system's accuracy stands at 91.92%. While this may appear to be low in comparison to stateof-the-art federated learning models, our work is noteworthy because, unlike others, it was achieved in a malicious vehicle setting. Despite the challenging environment, our method maintains high accuracy, making it a competent solution for preserving data privacy in vehicular networks.

*Index Terms*—Federated Learning, Blockchain, Vanet

## I. INTRODUCTION

AI-driven autonomous vehicles are gaining popularity due to their automation capabilities, allowing them to respond faster to road hazards than human drivers, potentially reducing traffic accidents caused by human error. They offer constant monitoring, lightning-fast responses, and precise navigation, making them superior to human drivers. Additionally, they hold promise as a reliable and convenient mode of transportation for elderly and disabled individuals.

Autonomous vehicles generate substantial amounts of data, encompassing sensor data, vehicle diagnostics, and real-time traffic information. Distributing this data among vehicles, manufacturers, and service providers while ensuring privacy and security is a challenging task [1], [2]. In order to process massive amounts of data in real time while keeping the data private and unaltered, a standardized, safe, and efficient system is required [3], [4], [5]. To address this challenge, it is essential to have a collaboration between technology, industry, and regulatory bodies. Using blockchain technology to build a trustworthy and decentralized data-sharing network is one strategy gaining favor. This would allow for open, safe, and efficient data sharing between vehicles and the infrastructure they interact with. In addition, federated learning can be used so that vehicles can gain knowledge from one another and apply it to their driving to enhance real-time performance and decision-making.

This paper presents a federated learning model incorporating blockchain to address the data-sharing problem in vehicular networks and enhance security by excluding malicious vehicles. Federated learning enables collaborative model training without sharing raw data, ensuring privacy and reducing data leaks, while blockchain technology provides a secure, tamper-proof ledger for sharing and verifying data. Blockchain enables the validation mechanism to verify the authenticity of participating vehicles, ensuring that only trustworthy and authorized nodes contribute to the collaborative learning process. The overall contribution of this study includes a system based on Federated Learning and blockchain that offers a potential solution to data sharing and security issues in VANETs. Our proposed model:

- Creates a real-world scenario where vehicles in VANETs act as workers, validators, or miners, contributing to collaborative model training.
- Employs a validation mechanism to validate and filter out distorted model updates from malicious nodes, ensuring the integrity of shared data.

The paper demonstrates the effectiveness of the validation mechanism in maintaining a consistently high accuracy even in the presence of malicious nodes, showcasing the model's resilience to adversarial attacks.

The rest of the paper is organized as follows: Section II presents a literature review on (autonomous) vehicle data sharing and approaches for solving relevant issues. Section III explains our proposed architecture in detail. Section V briefly discusses the proposed architecture's implementation and evaluation. Section VI concludes the work presented here, including the direction of the future work.

## II. RELATED WORK

A privacy-preserving approach [6] for vehicular networks, combining Federated Learning and blockchain, resulted in a 7.1% accuracy reduction compared to centralized machine learning. The expansion of autonomous driving features in intelligent vehicles raises concerns about potential safety risks, with the possibility of security breaches by installing malicious software. Using a Blockchain-based Internet of Things (IoT) system, Jabbar et al. [7] address the safety concerns related to autonomous driving features in self-driving vehicles. The proposed system aims to establish a decentralized cloud computing platform and ensure secure communication, thus effectively overcoming common challenges. The approach underwent rigorous testing to evaluate its performance and resilience in the face of security threats, successfully resolving key issues in Vehicle-to-Everything (V2X) communications. Pokhrel et al. [8] developed a blockchain-based federated learning (BFL) for private vehicular communication, optimizing the block arrival rate (the frequency at which new blocks are added to a blockchain) mathematically but lacking consideration of security attacks. Otoum et al. [9] proposed a Federated Learning and Blockchain solution with 97% accuracy, improved data privacy, and network security. Their solution, however, lacked sufficient scalability to validate malicious devices. In [10], the authors integrated permissioned blockchain with privacy-preserved federated learning for datasharing in industrial IoT, achieving high accuracy, efficiency, and enhanced security. Yang et al. [11] developed a trust management system for vehicular networks that is decentralized and uses blockchain technology. However, their focus was on trust computation and they did not address the issue of privacy preservation. In another study [12], they proposed a low-computational Directed Acyclic Graph (DAG) that utilizes blockchain technology to model the transaction relationships within the dynamic Internet of Vehicles. In contrast, Singh et al. [13] proposed a blockchain-based framework for intelligent vehicles, handling authentication, trust, and dynamic traffic rates, with intellectual vehicle trust point (IVTP) evaluating the real-time dependability of alternative transportation modes. Liu et al. [14] proposed a blockchain-based cloud and edge computing architecture for electric vehicles, enabling cooperative data sensing, energy sharing, and information analysis. Sey et al. [15] proposed a blockchain-based data protection architecture for IoV networks, though its lack of scalability and compatibility may prevent its widespread usage. Singh et al. [16] analyzed smart car privacy preservation but neglected in-vehicle network security and sensor reliability for communication. Calandriello et al. [17] focused on enhancing frame robustness in VANETs through alias employment with tone instruments while maintaining security and privacy. Memon et al. [18] used the SUMO simulator to generate dynamic aliases for blend-zone landscapes, improving vehicle location privacy and delicacy. Ali et al. [19] proposed a blockchainbased authentication method for VANETs ensuring privacy and transparency during signature verification of fake units. Su et al. [20] used a computationally costly proof-of-work consensus technique to solve traditional IoV's primary problem with two-way authentication and key agreement. Malik et al. [21] presented a VANET mutual authentication method employing a proprietary blockchain, distributed ledger access control, and certificate revocation to combat hostile nodes. Lu et al. [22] introduced a privacy-preserving federated learning technique for data privacy in Vehicle Cloud Computing Systems (VCPS), with the assumption of all parties being honest, a potentially unrealistic scenario. Li et al. [23] presented a data-sharing scheme for vehicular services utilizing an AI-enabled MultiAccess Edge Computing (MEC) server to achieve lower latency compared to centralized methods. However, the practical applicability of these approaches requires validation through real-world experiments. Zhou et al. [24] introduced a two-layer federated learning framework aimed at improving learning in 6G vehicular networks, but its applicability may be constrained by the necessity for a 6G network. In a different approach, Joshi et al. [25] presented ROAC-B, a clustering algorithm, and a blockchain-based data transmission technique designed for secure data sharing in cluster-based VANETs.

Following an extensive review of the existing literature, it was observed that most studies utilize technologies such as blockchain and federated learning to overcome the challenges faced in vehicular networks. However, this paper also takes a similar approach but utilizes various techniques, such as a dynamic role-based system that allows devices to adapt to specific roles like worker, validator, or miner. The validation mechanism identifies and filters out malicious updates, thus improving data security and integrity. Furthermore, the model's dynamic dataset distribution and role-switching capabilities provide unmatched flexibility that aligns well with the dynamic nature of vehicular networks.

## III. OVERVIEW OF THE PROPOSED SYSTEM ARCHITECTURE

Our system architecture is presented in Fig.1. Each node vehicle will have a local model that will train on its own data and then send updates to the validator. If the validator detects malicious data, they will send it to the nearest miner with a negative vote; otherwise, they will transmit it with a positive vote. A block will be generated using the data of the miner with the most positively voted data among all miners. The blockchain compares the new block to the previous one. If they match, it broadcasts to all vehicles; otherwise, the block is destroyed. Initial experiments demonstrate that, in comparison to conventional decentralized federated learning, the suggested solution can greatly increase the performance and security of vehicular networks

## IV. IMPLIMENTATION & METHODOLOGY

### A. Data Preprocessing

The MNIST dataset [26] is often used to evaluate different Machine Learning implementations. This dataset is split over four files, each of which contains images and labels for use in training and evaluation purposes. Only images and labels were to be extracted from the dataset, so preprocessing was minimal. In order to encode the labels, we turned to one-hot coding vectorization. In each round, the participant vehicles receive a randomly distributed distinct data set that is considered a local dataset for each vehicle. This distribution brings the process closer to a real-world application.

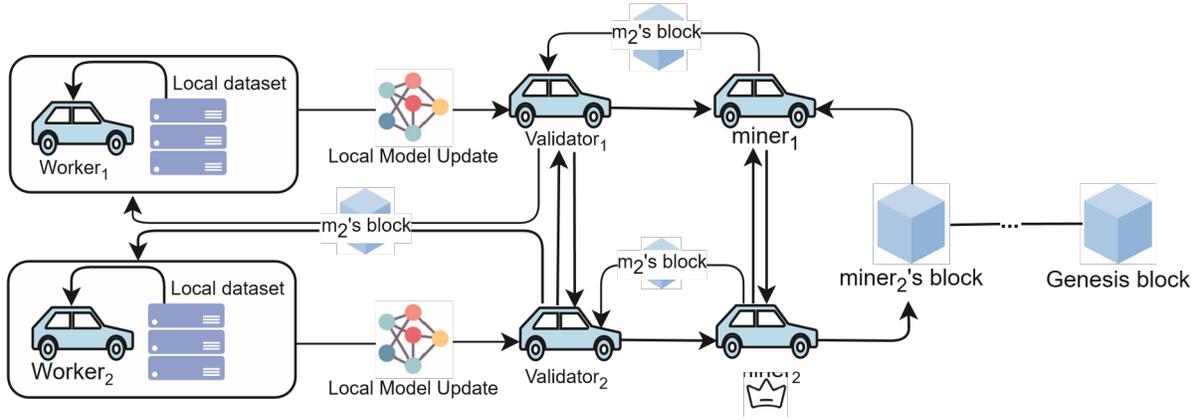

Fig. 1: Data Sharing using Blockchain Enabled Federated Learning for Interconnected Vehicular Network

### B. Training Process

Vehicles are seemingly classified as "worker," "validator," or "miner" based on the arguments provided. Each device with its own dataset is randomly assigned one of three roles during each round of communication:

- A work vehicle is undergoing local model training;
- A validator casts votes on the validity of the received local model parameters; or
- A miner integrates voting outcomes with the corresponding local model parameters into the subsequent consensual block.

A CNN model is used to train on the local dataset owned by the vehicles. In the background of the training model are 2D convolution layers. All instances of Kernel were set to 2, whereas stride varied between 1 in convolution layers and 2 in max-pooling layers.

### C. Operations of The Proposed Model

In the same way as Vanilla FL [27] learns through a series of communication rounds, the model is controlled by a set of vehicles ($V = v_1, v_2, .., v_n$). Each vehicle participates in each round by acting as either a worker, validator, or miner, depending on whatever job it is allocated. A vehicle's public key serves as its id and can be used to validate the authenticity of any blocks or transactions it creates. In the communication round ($c_i$), a worker updates their local model by performing local learning on the global model ($g_{i-1}$) they built in the previous round ($c_{i-1}$). Local model updates would be validated before being used to build $g_i$. Additionally, the worker vehicle derives its anticipated rewards for learning by adhering to the reward mechanism. The worker vehicle encapsulates both local model updates and rewards in a transaction denoted as $t_i$, and then sends it to a randomly associated validator. Each validator receives $t_i$ from all of the workers with which it is affiliated and then broadcasts $t_i$ to the rest of the validators. Worker vehicles receive a reward when the validator positively votes on them. The validator receives a verification reward for verifying one $t_i$. If $t_i$ is a verified transaction, the validator will take the local model update from $t_i$ and vote on it (positive or negative) according to the voting mechanis. A Validator is also rewarded for validation by voting on a single local model update. The validator's vote on the local model update, along with the verification reward and validation reward, will be included in a validator transaction ($t_v$) that is signed by the validator's private key and sent to the associated miner. A miner broadcasts the received $t_v$ to the rest of the miners. By doing so each miner would have access to every voting result. The authentication verification of a single $t_v$ is worth a reward to the miner. After the signature is validated, the validator's vote will be extracted from the $t_v$. For a local update to a parameter, the miner will compile the votes from all of the validators that have already been extracted. The results of each worker vehicle's vote are combined and then added to a privately built candidate block. Workers' vehicles', validators', and miners' prospective rewards would all be listed in the candidate block. Following the consensus, the miner proceeds to mine their own candidate block. To do this, they hash the entire block and sign the resulting hash with their private key. Once the miner has successfully mined a block, they distribute it to all the other miners within the network. Once all the mined blocks have been received from the network, the miner then selects the miner-generated block with the highest rewards, as the legitimate block. The reward record and voting results with related model updates can be retrieved from only this legitimate block. This valid block would be added by each miner to their respective blockchain and then request that the worker vehicle and validator linked with it also download this block to add to their own blockchains. To account for the possibility that certain propagated blocks might suffer a prolonged transmission delay, so, in order for miners to accept the forwarded blocks, a time limit should be imposed. In addition, if a local model update obtains more Negative votes than Positive votes, the related worker is marked as possibly malicious, and hence no rewards are given to this worker. Each vehicle should blacklist a worker and refuse any future communications from it if it is identified as malicious for several consecutive communication rounds. The proposed model employs a role-switching strategy in each new round, reducing the likelihood of a consistently malicious miner or validator. Miners and validators are re-selected in each new round, which helps mitigate potential issues with malicious entities consistently influencing the system. If a vehicle is identified as malicious for several consecutive communication rounds, it is blacklisted. This means that other vehicles refuse any future communications from the blacklisted vehicle, whether those are workers, validators, or miners, thereby adding an extra layer of security to the system.

### D. Validator Voting Mechanism

In order to prevent severely distorted local model updates provided by malicious worker vehicles, the Validator Voting Mechanism was developed. In the experiments, we seeded the network with 20 vehicles, 17 of which were worker vehicles and 3 of which were malicious. In order to evaluate the efficacy of each vehicle, we reallocated their roles in each communication round. We maintained a consistent role combination for devices, ensuring a constant presence of 12 worker vehicles, 5 validators, and 3 miners to maintain an adequate number of worker vehicles and validators. The MNIST training set was randomly shared among all 20 vehicles, and the MNIST test set was used by each vehicle as its own local test set.

Algorithm 1 presents the proposed validation mechanism for local model updates.

---

**Algorithm 1 Validation Mechanism**

1: Input:
2: $L_W$ = Locally updated model by worker vehicle;
3: $A_v$ = Accuracy of the Locally updated model evaluated by validator using $Test_v$.
4: $Test_v$ = Test set of validators;
5: $V_h$ = The tolerance for accuracy-drop measurement between $A_v$ and $A_w$.
6: Initialize:
7: $V_T$ = vote by the validator on local model update;
8: $g_i$ = Constructed global model at a communication round 9: $A_w$ = Accuracy of the Locally updated model using $Test_w$.
10: $R_{vali}$ = validation-reward
11: Evaluate by the validator ( $L_W, Test_v$ ) return $R_{vali}$ , $A_w$
12: **if** $A_v - A_w > V_h$ **then**
13:     VT = Negative;
14: else
15:     VT = Positive; ;
16: end if
17: return $R_{vali}$ , $V_T$

---

### E. Reward Mechanism

To ensure the security of blockchain-based federated learning, it is essential to prevent the selection of a block produced by a malicious vehicle, as a malicious miner could disrupt the computation of the global model by including fraudulent voting results and validator signatures in a mined block used to record voting results. For this reason, the proposed model provides vehicles with rewards based on their roles, which is influenced by the reward mechanism in BlockFL [28] and strengthened by the role-switching policy.

*1) Rewards Mechanism for Worker:* In a communication round, a worker vehicle reward is proportional to the number of data samples in the training set of that worker vehicle and the number of training epochs in that communication round. Suppose in a communication round from the validator, $V_P$ = The count of positive votes for the locally updated model. $V_n$ = The count of Negative votes for the locally updated model.
In that case, the total rewards for a worker in a communication round are calculated as follows: if $A_v - A_w > V_h$ then $V_T$ = Negative; else

$V_T$ = Positive; end
if

*2) Rewards Mechanism for Validator:* In each communication round, a validator is incentivized for authenticating signatures of received worker-transactions ($t_i$) and endorsing a locally updated model derived from the verified $t_i$. The total rewards for a validator in a communication round are calculated as follows:
$r_{veri}$ = Validator's verification-reward, by verifying the signature of one worker-transaction.
$r_{vali}$ = Validator's validation-reward, by voting on one locally updated model. r = voting unit.

$$reward = |r_{veri}| + |r_{vali}| = |t_i| \times r + |t_v| \times r - - -(i)$$

### F. Rewards Mechanism for Miner

In a communication round, the miner is rewarded for the verification of signatures associated with validatortransactions, $t_v$, calculated as:

$$Reward = t_v \times \alpha - - -(ii)$$

### G. Miner Selection

Worker vehicles contribute the most to the global model's local learning, making them the most important part of the training process. They are rewarded accordingly based on the cumulative blocks they generate. The block selection for the global model update is determined by the miner with the highest positive vote, signifying the most legitimate contribution to the learning process. This chosen block becomes the legitimate one for the current round, and the miner behind it is called the winning miner. To prevent potential issues with malicious vehicles, the role-switching strategy ensures that miners are re-selected in each new round, reducing the likelihood of a malicious vehicle consistently becoming a miner. The "nondemocracy side effect" is also mitigated, as it avoids always choosing the malicious device with the most affirmative votes as the winning miner. This helps safeguard the learning process from potential attacks. Any miner's ID appearing on a blacklist leads to a refusal by vehicles to accept any blocks from that miner. This measure adds a layer of security to the system.

## V. RESULTS AND DISSCUSSION

### A. Experimental Results

We conducted all the experiments in the Colab cloud machine where the resources were 12GB of RAM and 16 GB of Graphics Memory. The experiments were run for almost 4 hours each. The runtime depends on the parameters given and the capability of the resources. We ran two experiments on this system, They are listed below:

- Network of 20 vehicles and 0 of them were malicious with a validation threshold of 1.00 since there was nothing to validate.
- Network of 20 vehicles, where 3 of them would be malicious, with a validation threshold of 0.08. In both cases, there were 100 rounds of communication performed, with an epoch of 5, a learning rate of 0.01, and a batch size of 10 in each round for a vehicle.

### B. Effectiveness of Validation Mechanism

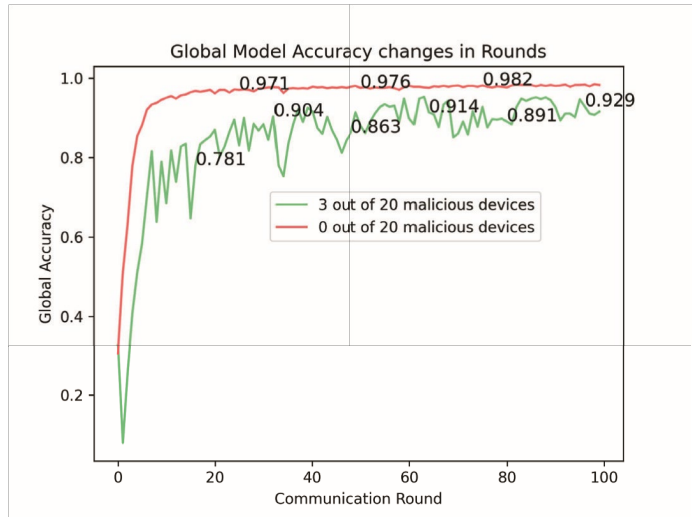

Fig. 2: Global Accuracy changes in our model, with and without malicious vehicles

Fig. 2 demonstrates the efficacy of the proposed validation mechanism. We conducted two executions comprising 100 communication rounds for each of the experiments indicated in the figure. The accuracy of the global model was recorded at the end of each round based on data from 20 devices. In all executions across all 100 rounds, we manually assigned a combination of 12 worker vehicles, 5 validators, and 3 miners. For both scenarios, where each vehicle utilized the entire MNIST dataset, and the global model remained consistent across vehicles after each communication round, the evaluated accuracy is depicted in the figure. The system yields a considerably high accuracy in the absence of any malicious vehicles. However, even with three malicious vehicles, the accuracy experiences a minor decline. Although the system encounters slightly more challenges, it maintains a consistent accuracy comparable to the scenario without maliciousness.

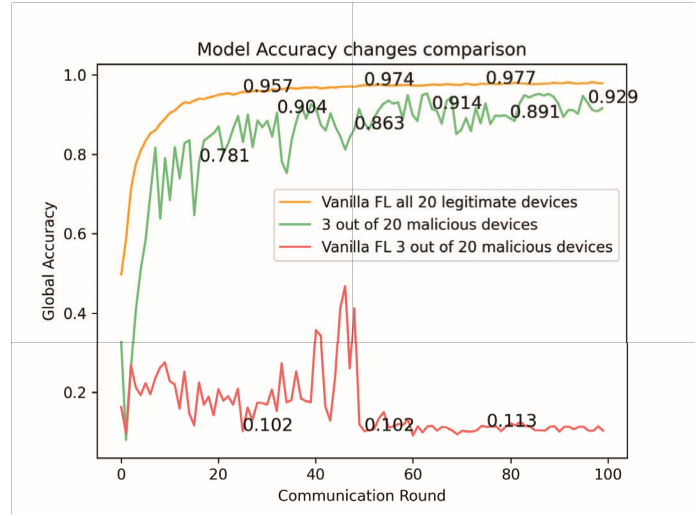

Fig. 3: Accuracy Comparison with Vanilla FL with and without malicious vehicles

The accuracy of Vanilla Federated Learning is depicted by the orange curve area, which encompasses all 20 legitimate vehicles. On the other hand, the proposed model's accuracy, with 3 out of 20 malicious participants, is represented by the green curve area in Fig. 3. Lastly, the red curve area illustrates the accuracy of Vanilla FL when confronted with an equal number of malicious participants.

### C. Vanilla Federated Learning with malicious vehicles

The red curve area illustrates the accuracy of Vanilla FL when confronted with malicious participants. It's worth noting that the Vanilla federated learning process's accuracy drastically decreases when malicious vehicles participate. During the model update exchanges, malicious vehicles can inject explicit noise or biased data, which can lead to inaccurate results. It's possible to intentionally mislead the learning process and undermine the robustness of the global model in this way. These malicious vehicles attempt to disrupt the convergence of the model and slow down the learning process by injecting incorrect or misleading gradients. The malicious updates can be designed to resemble legitimate updates, making it difficult to detect and filter out harmful contributions. The inaccurate data from the malicious vehicles might therefore have a major impact on the global model, leading to a drastic reduction in accuracy.

TABLE I: PERFORMANCE ANALYSIS COMPARISON

| | | Accuracy | |
|---|---|---|---|
| Reference | Focused Method | Including Malicious Node | No Malicious Node |
| Moulahi et al. [6] | SVM+ Federated Learning | × | 82.45% |
| Kakkar et al. [29] | Random Forest + Blockchain | × | 93% |
| Otoum et al. [9] | Federated Learning + Blockchain | × | 97% |
| Rajan et al. [30] | Multi-layered federated extreme learning + Blockchain | × | 98% |
| Chen et al. [31] | Federated Learning + Blockchain + Validation | 87% | × |
| Proposed | Federated Learning + Blockchain + Validation | 98.99% | 91.92% |

*D. Proposed model with malicious vehicles*

With 3 out of 20 malicious participants, the proposed model's accuracy is depicted by the green curve area in Fig. 2. The validation mechanism effectively filters out distorted local model updates from malicious devices, resulting in an overall accuracy that approaches the green curve. However, the orange and green curves have a roughly 10% gap after convergence due to inadequate learning of the three malicious devices that own up to 15% of the training data. The validation mechanism serves as a protective measure against potentially harmful vehicles. The server can detect anomalies in the performance of individual vehicles by evaluating the updated model parameters. If a vehicle constantly sends malicious updates, its validation accuracy will differ significantly from that of the other vehicles that behave properly. The vehicles that deviate from the norm can be identified, and their updates can either be discarded or given less weight during the aggregation process. We utilize the federated averaging (FedAvg) technique, which involves averaging the model updates from multiple vehicles. These techniques prioritize vehicles that exhibit consistent and reliable performance, placing greater importance on their behavior. By assigning less importance or ignoring the updates from malicious vehicles, we can reduce the impact of their potentially harmful contributions.

*E. Performance Analysis Comparison*

Table I presents a comprehensive overview of the results from prior research, depicting the accuracy of visualization. The four mentioned models share common characteristics with the proposed model, making them suitable candidates for comparison. One notable similarity is that all models are decentralized and rely on distributed computing and decisionmaking. Moulahi et al. [6] and Kakkar et al. [24] employed machine learning and Federated Learning, achieving 82.45% and 93% accuracy, respectively. However, their decentralized methods lacked consideration for the presence of malicious nodes, which is crucial for real-world robustness. Similarly, Otoum et al. [9] and Rajan et al. [30] integrated Federated Learning and Blockchain, achieving notable accuracies of 97% and 98%, respectively. Yet, their robustness under adversarial conditions remained unclear due to limited impact analysis of malicious nodes.

Malicious nodes can significantly reduce the accuracy of these models by injecting incorrect, biased, or noisy data, distorting model updates, and compromising overall performance. To address this challenge, the proposed model incorporates a validation mechanism, ensuring accuracy maintenance even with malicious participants, achieving 91.92% accuracy in their presence and 98.99% without, surpassing state-of-theart decentralized models' resilience to attacks. Furthermore, compared to Chen et al.'s [31] approach, which achieved 87% accuracy, the proposed model's efficiency lies in its communication around process optimization, enabling streamlined model updates exchange, verification, and aggregation. The proposed model's dynamic dataset and role switching for vehicles maintain consistency in the presence of malicious entities, offering a significant advantage over Chen et al.'s [31] fixed dataset approach, which is more aligned with reallife scenarios. Demonstrated in Fig. 2, the proposed model showcases remarkable accuracy retention, even with malicious vehicles, in contrast to the observed decrease in accuracy with Vanilla Federated Learning. In contrast to the observed decrease in accuracy with malicious vehicles in Vanilla Federated Learning, illustrated in Fig. 2, our approach maintains accuracy levels close to those without malicious vehicles, which is impressive.

*F. Discussion*

The proposed model achieves an accuracy level of 98.99% in the absence of malicious nodes; even when malicious nodes are present, it still maintains an impressive accuracy of 91.92%. The validation mechanism introduced in the model plays a significant role in maintaining accuracy by filtering out any distorted model updates from malicious nodes. The role-based system, which assigns roles to vehicles (workers, validators, miners), also helps to enhance the overall security of the model by minimizing the impact of malicious miners and validators.

Compared to traditional decentralized federated learning techniques, the proposed approach integrating Blockchain and Federated Learning ensures data privacy and security, safeguarding sensitive information and making it an ideal solution for preserving data privacy in vehicular networks. Nevertheless, there are certain limitations associated with the proposed approach. Scaling up the network may introduce additional challenges that our experiments with a limited number of vehicles still need to address. Additionally, the impact of network delays on model updates and the potential need for latency mitigation strategies should be considered.

## VI. CONCLUSION AND FUTURE WORK

This study demonstrates the development of a decentralised and federated learning network through the utilisation of blockchain technology. The primary objective is to augment the security of data-sharing among autonomous vehicles. The proposed model also demonstrates the efficacy of validation in the context of a decentralised model, wherein real-life contributors can be present. It is reasonable to anticipate that only a subset of contributors will exhibit perfection and provide appropriate input. The implementation of validation and decentralised voting mechanisms serves to maintain consistency and accuracy in the global model's training process by providing protection against malicious or tampered vehicles. The model presented in our study is not merely a training algorithm, but rather a network configuration strategy designed to train models while preserving data privacy and maintaining consistent global accuracy, even in challenging conditions.

Moving forward, we plan to refine the algorithm, improve concurrent processing, and better manage threads. A future focus will also be the inclusion of a dedicated data quality assessment during both pre-processing and model validation, further enabling the distinction between malicious inputs and genuine data quality issues.

REFERENCES

[1] J. Joy and M. Gerla, "Internet of vehicles and autonomous connected car- privacy and security issues," in *2017 26th International Conference on Computer Communication and Networks (ICCCN)*. IEEE, 2017, pp. 1–9.
[2] K. F. Hasan, A. Overall, K. Ansari, G. Ramachandran, and R. Jurdak, "Security, privacy, and trust of emerging intelligent transportation: cognitive internet of vehicles," in *Next-Generation Enterprise Security and Governance*. CRC Press, 2022, pp. 193–226.
[3] S. Zhang, M. Lagutkina, K. O. Akpinar, and M. Akpinar, "Improving performance and data transmission security in vanets," *Computer Communications*, vol. 180, pp. 126–133, 2021.
[4] K. Ansari and K. F. Hasan, "Proposition of augmenting v2x roadside unit to enhance cooperative awareness of heterogeneously connected road users," *arXiv preprint arXiv:2305.14809*, 2023.
[5] M. R. Al Asif, K. F. Hasan, M. Z. Islam, and R. Khondoker, "Stridebased cyber security threat modeling for iot-enabled precision agriculture systems," in *2021 3rd International Conference on Sustainable Technologies for Industry 4.0 (STI)*. IEEE, 2021, pp. 1–6.
[6] T. Moulahi, R. Jabbar, A. Alabdulatif, S. Abbas, S. El Khediri, S. Zidi, and M. Rizwan, "Privacy-preserving federated learning cyber-threat detection for intelligent transport systems with blockchain-based security," *Expert Systems*, p. e13103, 2022.
[7] R. Jabbar, M. Kharbeche, K. Al-Khalifa, M. Krichen, and K. Barkaoui, "Blockchain for the internet of vehicles: A decentralized iot solution for vehicles communication using ethereum," *Sensors*, vol. 20, no. 14, p. 3928, 2020.
[8] S. R. Pokhrel and J. Choi, "Federated learning with blockchain for autonomous vehicles: Analysis and design challenges," *IEEE Transactions on Communications*, vol. 68, no. 8, pp. 4734–4746, 2020.
[9] S. Otoum, I. Al Ridhawi, and H. T. Mouftah, "Blockchain-supported federated learning for trustworthy vehicular networks," in *GLOBECOM 2020-2020 IEEE Global Communications Conference*. IEEE, 2020, pp. 1–6.
[10] Y. Lu, X. Huang, K. Zhang, S. Maharjan, and Y. Zhang, "Blockchain empowered asynchronous federated learning for secure data sharing in internet of vehicles," *IEEE Transactions on Vehicular Technology*, vol. 69, no. 4, pp. 4298–4311, 2020.
[11] Z. Yang, K. Yang, L. Lei, K. Zheng, and V. C. Leung, "Blockchain-based decentralized trust management in vehicular networks," *IEEE internet of things journal*, vol. 6, no. 2, pp. 1495–1505, 2018.
[12] W. Yang, X. Dai, J. Xiao, and H. Jin, "Ldv: A lightweight dagbased blockchain for vehicular social networks," *IEEE Transactions on Vehicular Technology*, vol. 69, no. 6, pp. 5749–5759, 2020.
[13] M. Singh and S. Kim, "Branch based blockchain technology in intelligent vehicle," *Computer Networks*, vol. 145, pp. 219–231, 2018.
[14] C. Huang, R. Lu, X. Lin, and X. Shen, "Secure automated valet parking: A privacy-preserving reservation scheme for autonomous vehicles," *IEEE Transactions on Vehicular Technology*, vol. 67, no. 11, pp. 11169–11180, 2018.
[15] C. Sey, H. Lei, W. Qian, X. Li, L. D. Fiasam, S. L. Kodjiku, I. AdjeiMensah, and I. O. Agyemang, "Vblock: A blockchain-based tamperproofing data protection model for internet of vehicle networks," *Sensors*, vol. 22, no. 20, p. 8083, 2022.
[16] M. Singh and S. Kim, "Blockchain based intelligent vehicle data sharing framework," *arXiv preprint arXiv:1708.09721*, 2017.
[17] J. Huang, M. Zhao, Y. Zhou, and C.-C. Xing, "In-vehicle networking: Protocols, challenges, and solutions," *IEEE Network*, vol. 33, no. 1, pp. 92–98, 2018.
[18] G. Calandriello, P. Papadimitratos, J.-P. Hubaux, and A. Lioy, "Efficient and robust pseudonymous authentication in vanet," in *Proceedings of the fourth ACM international workshop on Vehicular ad hoc networks*, 2007, pp. 19–28.
[19] I. Memon, Q. Ali, A. Zubedi, and F. A. Mangi, "Dpmm: dynamic pseudonym-based multiple mix-zones generation for mobile traveler," *Multimedia Tools and Applications*, vol. 76, pp. 24359–24388, 2017.
[20] I. Ali, M. Gervais, E. Ahene, and F. Li, "A blockchain-based certificateless public key signature scheme for vehicle-to-infrastructure communication in vanets," *Journal of Systems Architecture*, vol. 99, p. 101636, 2019.
[21] N. Malik, P. Nanda, A. Arora, X. He, and D. Puthal, "Blockchain based secured identity authentication and expeditious revocation framework for vehicular networks," in *2018 17th IEEE International Conference On Trust, Security And Privacy In Computing And Communications/12th IEEE International Conference On Big Data Science And Engineering (TrustCom/BigDataSE)*. IEEE, 2018, pp. 674–679.
[22] Y. Lu, X. Huang, Y. Dai, S. Maharjan, and Y. Zhang, "Federated learning for data privacy preservation in vehicular cyber-physical systems," *IEEE Network*, vol. 34, no. 3, pp. 50–56, 2020.
[23] X. Li, L. Cheng, C. Sun, K.-Y. Lam, X. Wang, and F. Li, "Federatedlearning-empowered collaborative data sharing for vehicular edge networks," *IEEE Network*, vol. 35, no. 3, pp. 116–124, 2021.
[24] X. Zhou, W. Liang, J. She, Z. Yan, I. Kevin, and K. Wang, "Twolayer federated learning with heterogeneous model aggregation for 6g supported internet of vehicles," *IEEE Transactions on Vehicular Technology*, vol. 70, no. 6, pp. 5308–5317, 2021.
[25] G. P. Joshi, E. Perumal, K. Shankar, U. Tariq, T. Ahmad, and A. Ibrahim, "Toward blockchain-enabled privacy-preserving data transmission in cluster-based vehicular networks," *Electronics*, vol. 9, no. 9, p. 1358, 2020.
[26] "MNIST Dataset," https://www.kaggle.com/datasets/hojjatk/ mnist-dataset, [Accessed 27-07-2023].
[27] S. R. Pokhrel and J. Choi, "A decentralized federated learning approach for connected autonomous vehicles," in *2020 IEEE Wireless Communications and Networking Conference Workshops (WCNCW)*. IEEE, 2020, pp. 1–6.
[28] H. Kim, J. Park, M. Bennis, and S.-L. Kim, "Blockchained on-device federated learning," *IEEE Communications Letters*, vol. 24, no. 6, pp. 1279–1283, 2019.
[29] D. Rajan, P. Eswaran, G. Srivastava, K. Ramana, and C. Iwendi, "Blockchain-based multi-layered federated extreme learning networks in connected vehicles," *Expert Systems*, p. e13222.
[30] H. Chai, S. Leng, Y. Chen, and K. Zhang, "A hierarchical blockchainenabled federated learning algorithm for knowledge sharing in internet of vehicles," *IEEE Transactions on Intelligent Transportation Systems*, vol. 22, no. 7, pp. 3975–3986, 2020.
[31] H. Chen, S. A. Asif, J. Park, C.-C. Shen, and M. Bennis, "Robust blockchained federated learning with model validation and proof-ofstake inspired consensus," *arXiv preprint arXiv:2101.03300*, 2021.